\documentclass[fleqn,12pt]{wlscirep}
\usepackage[utf8]{inputenc}
\usepackage[T1]{fontenc}
\usepackage{tabularx}
\usepackage{pdfpages}

\title{LLMs with Personalities in Multi-issue Negotiation Games}

\author[1]{Sean Noh}
\author[1]{Ho-Chun Herbert Chang}
\affil[1]{Dartmouth College, Program in Quantitative Social Science, Hanover NH 03755, USA}

\affil[*]{herbert.chang@dartmouth.edu}

\affil[+]{these authors contributed equally to this work}


\begin{abstract}
Powered by large language models (LLMs), AI agents have become capable of many human tasks. Using the most canonical definitions of the Big Five personality, we measure the ability of LLMs to negotiate within a game-theoretical framework, as well as methodological challenges to measuring notions of fairness and risk. Simulations (n=1,500) for both single-issue and multi-issue negotiation reveal increase in domain complexity with asymmetric issue valuations improve agreement rates but decrease surplus from aggressive negotiation. Through gradient-boosted regression and shapley explainers, we find high openness, conscientiousness, and neuroticism are associated with fair tendencies; low agreeableness and low openness are associated with rational tendencies. Low conscientiousness is associated with high toxicity. These results indicate that LLMs may have built-in guardrails that default to fair behavior, but can be “jail broken” to exploit agreeable opponents. We also offer pragmatic insight in how negotiation bots can be designed, and a framework of assessing negotiation behavior based on game theory and computational social science. 
\end{abstract}
\begin{document}

\flushbottom
\maketitle
%
%
\thispagestyle{empty}

\section*{Introduction}

Powered by large language models (LLMs), AI agents have become capable of many human tasks. This presents the risk of agents exploiting humans, as well as methodological challenges to measuring notions of fairness and risk. While existing research has assessed the ability of agents to bargain and bid, there is a gap between the game theory of bargaining and linguistic analysis. Combining game theory with LLMs and natural language processing, this study aims to assess AI agents' ability to bargain, by using LLMs to simulate negotiations between agents with OCEAN-based personalities. 

Negotiation is a fundamental task for humans. It is a process where parties can settle issues, discover surplus, and create value. Due to it's ubiquity, it has been widely studied in many contexts, including economics~\cite{osborne1994course, raiffa1982art}, business~\cite{walsh1999modeling,lewicki2016essentials,ehlich2011discourse,huang2010agent}, communication~\cite{arvanitis2011negotiation,maaravi2011negotiation}, and behavioral psychology~\cite{rubin2013social,de2007psychology}. In the social sciences, a significant stream considers how individual differences contribute to divergent outcomes in negotiation. In particular, personality traits have been shown to modulate bargaining in a few different ways. The Big Five personality traits, particularly neuroticism, extraversion, and agreeableness, are associated with collaborative, compromising, or collaborative behavior~\cite{ma2005exploring,marlowe1966opponent}. Personality also has influences on ultimatum bargaining decisions~\cite{brandstatter2001personality}, motivational orientations such as optimism~\cite{rubin2013social}, and other typologies of negotiation predilections such as aggression and submission~\cite{nassiri2008personality}. However, other studies have shown situational factors may play a larger role, including the rhetoric used~\cite{morris1999misperceiving}.

In tandem, negotiation can also be formalized through the lens of game theory, specifically bargaining games. Canonically, a negotiation game is a two player game where players divide one or more issues, with each issue being valued differently by each agent. Each agent aims to maximize their utility. At the start of the game, agents are informed of the maximum rounds to split the issues. If no agreement is reached by the final round, both players receive nothing. Each round consists of an \textit{offer} and \textit{response}. An offer includes a proposed division of the issues. A response either accepts the offer, ending the game and dividing the issues accordingly, or rejects the offer, moving the game to the next round. On an agent’s turn, the agent responds to the previous offer and then, if rejecting the offer, makes a counteroffer. In the past decade, there has been a particular effort in artificial intelligence~\cite{chang2021multi, jennings2001automated, kraus1997negotiation, gerding2000scientific} to model autonomous agents capable of producing rational and fair outcomes.

The advent of ChatGPT in 2022 effectively combines these two streams of research. Since 2017, agents were shown to misrepresent their intentions within negotiation dialogues~\cite{lewis2017deal,chang2021multi}. In parallel, studies have not just proposed large-language models as a way to simulate subjects for social scientific experiments~\cite{yang2023harnessing}, but have also sought to quantify their ability to perform causal reasoning and rational decision making~\cite{liu2024large,duan2024gtbench}. 
For instance, it has been shown AI agents can "jailbroken" using persuasion techniques~\cite{zeng2024johnny}. We build on contemporary work considering personality and LLMs~\cite{mei2023turing} and negotiation on different base models~\cite{davidson2024evaluating}. Moreover, success in deep reinforcement learning (DRL) almost ten years ago in games such as Chess~\cite{silver2016mastering}, Go~\cite{silver2017mastering}, Poker~\cite{brown2018superhuman} and Atari games~\cite{mnih2013playing} have inspired DRL's application to complex human tasks, such as negotiation. Baarslag divides negotiation into three pillars--- the \textit{bidding strategy}, \textit{opponent modeling}, and \textit{acceptance strategy}. 

With LLMs at our disposal, the combination of both the quantitative and qualitative dimension of negotiation is now feasible.
Our research questions are as follows:
\begin{itemize}
    \item \textbf{RQ1:} Which personality-based agents yield the highest returns?
    \item \textbf{RQ2:} How does domain complexity influence payoffs for different agent types?
    \item \textbf{RQ3:} What language and rhetorical features are most prevalent amongst each of the agent-types?
    \item \textbf{RQ4:} What pairings head-to-head result in the highest level of exploitation?
    \item \textbf{RQ5:} Among agent-type, toxicity, round type, and rhetorical strategies, what yields the highest utilities?
\end{itemize}

\subsection*{Negotiation games}

To assess the ability of different personality agents to negotiate with each other, we created ten types of agents to compete against each other in both single- and multi-issue negotiations.  In each game, two agents were created as distinct instances of the \textit{gpt-4-turbo} model using the ChatGPT API. Each agent was defined as one of ten personality types: either a high or low level of \textit{openness}, \textit{conscientiousness}, \textit{extroversion}, \textit{agreeableness}, or \textit{neuroticism}. Agent personalities were initialized [or The system content of each agent was initialized] based on personality facets described by Howard and Howard~\cite{howard1995big}.

Each agent played against all other agents, including itself, as both the first player to make an offer (P1) and the first player to respond to an offer (P2). For single-issue negotiations, agents divided $ \$100 $. For multi-issue negotiations, agents divided 10 apples, 10 bananas, and 10 crepes, which they valued according to a preference profile. P1 valued apples at $ \$1 $, bananas at $ \$2 $, and crepes at $ \$3 $. P2 had an opposite valuation of apples at $ \$3 $, bananas at $ \$2 $, and crepes at $ \$1 $. Agents took turns responding to and making offers until an agreement was reached or round six ended. If no agreement was made by the sixth round, the game ended in default and both agents received a payoff of $ \$0 $.

The purpose of this experiment is to analyze how effective different LLM-agent personalities are at negotiating with a deadline. Thus, it is important to consider the negotiations from a game theoretic perspective. In a six round game between two rational agents, P2 has a massive advantage as it is always able to make the final offer. P2 will decline all proposals until the final round, where it will offer the minimum to P1. P1 will accept any offer with a greater payoff than $ \$0 $ because if it declines, the game will end in default and P1 will receive nothing. Thus, the Nash Equilibrium, or the outcome assuming perfect play between two rational agents, is a $ \$99 $ payoff for P2 and a $ \$1 $ payoff for P1 in the single-issue case. In the multi-issue case, P2 will give away one crepe, which it values least, resulting in an outcome of 10 apples, 10 bananas, and 9 crepes for a payoff of $ \$59 $. P1 will receive a payoff of $ \$3 $ for 1 crepe.

\section*{Results}

We ran ten trials of single-issue negotiation games and five trials of multi-issue games. Each trial consisted of all ten agent personalities playing against each other as both P1 and P2 for a total of 100 games per trial. This yields a total sample size of 1,000 single-issue games and 500 multi-issue games.

\subsection*{Agreeableness and Domain Complexity}

\begin{figure}[!htp]
\centering
\includegraphics[width=1.0\linewidth]{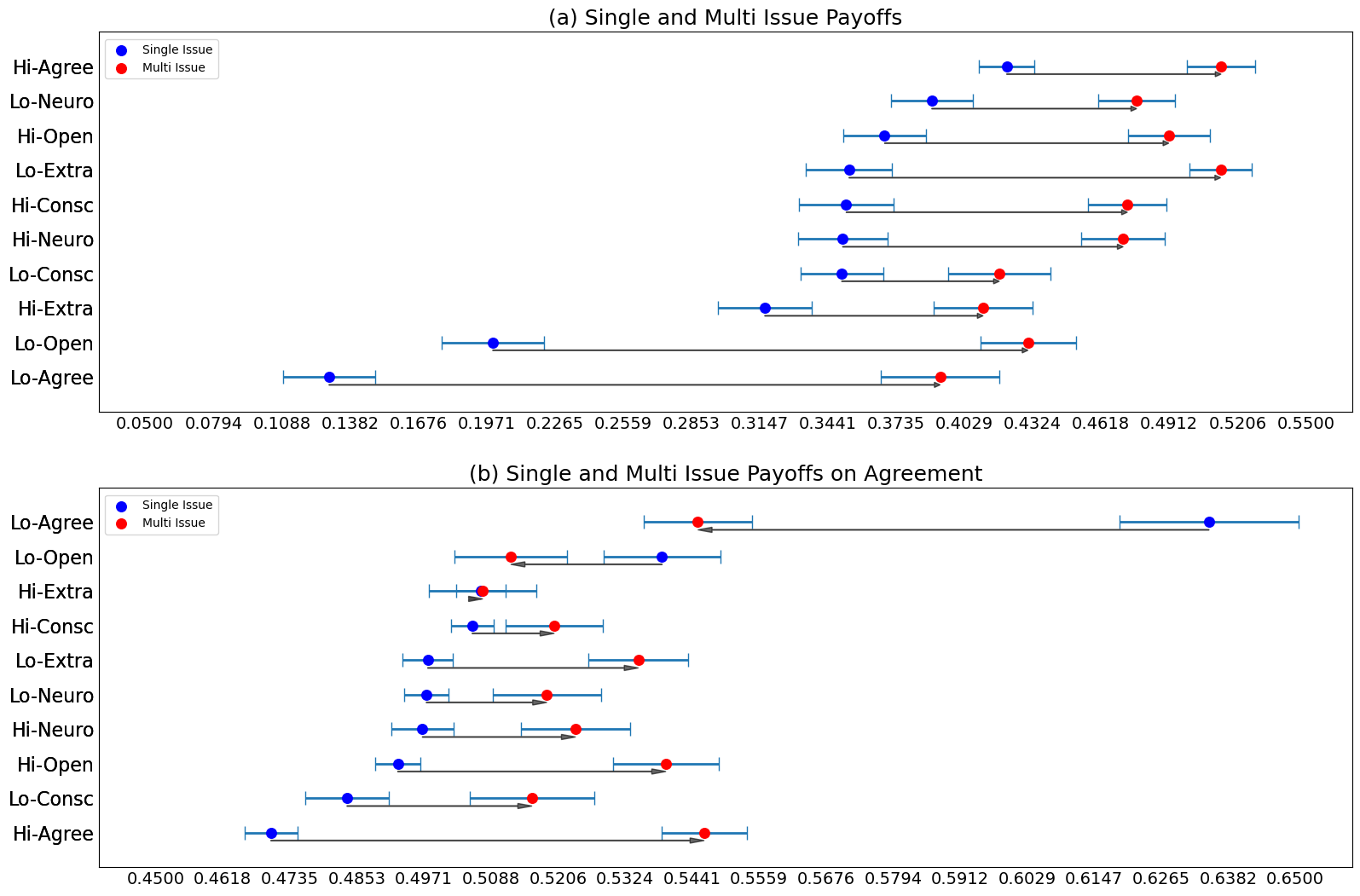}
\caption{Normalized payoffs of personality-based agents in single and multi issue games (a) including and (b) excluding games ending in default.}
\label{fig:payoffs}
\end{figure}

Figure~\ref{fig:payoffs} shows the normalized payoff of each agent personality in single and multi issue games, with and without games ending in default. We find that agreeableness influences average payoff the most. In both the single and multi issue games, the personalities with the highest payoffs had the lowest default rates (a comparison of default rates and payoffs can be found in the appendix). Based on the literature, these personalities tended to be more accommodating, more passive, and less ambitious, with personality facets such as "willing to help others", "open to reexamining values", "slow to anger", and "stays in background". However, while these personalities were able to reach more agreements, this does not characterize the full range of dynamics.

Conditional on an agreement being made, low agreeableness leads to high returns. In the single-issue case, Figure~\ref{fig:payoffs}b shows that low agreeableness and low openness, the personalities with the highest default rates, generated the highest value when an agreement was made. This indicates that agreeableness directly modulates the reserve price, or lowest payoff that an agent will accept. In the multi-issue case, while low agreeableness still had one of the highest returns, it did not have as extreme of an advantage over the other personalities. In fact, high agreeableness and high openness had a very similar average payoff. This is because in multi-issue games, these more cooperative personalities were best able to use the agents' differing preference profiles to find more efficient and lucrative deals.

Overall, we find that domain complexity with asymmetric issue valuations tends to increase payoffs for agents. Figure~\ref{fig:payoffs}a shows that all agents achieved higher payoffs in the multi-issue negotiations than the single-issue negotiations. One reason for this is a lower default rate in multi-issue negotiations. While agents came to an agreement in only 64.3\% of single-issue games, they reached agreement in 89.0\% of games in the multi-issue case. However, the increased payoff average in multi-issue games was not solely caused by more frequent agreements; Figure~\ref{fig:payoffs}b shows that even with defaulted games removed, almost all agents had a higher payoff in multi-issue games. This is because in multi-issue games, the different preferences of issues allows agents to find synergistic trade-offs that increase both agents' utilities. Multi-issue games allow for better deals, which results in more agreements. Together, these answer \textbf{RQ1} and \textbf{RQ2}.

\subsection*{Agent-based differences}

\begin{figure}[htp]
\centering
\includegraphics[width=1.0\linewidth]{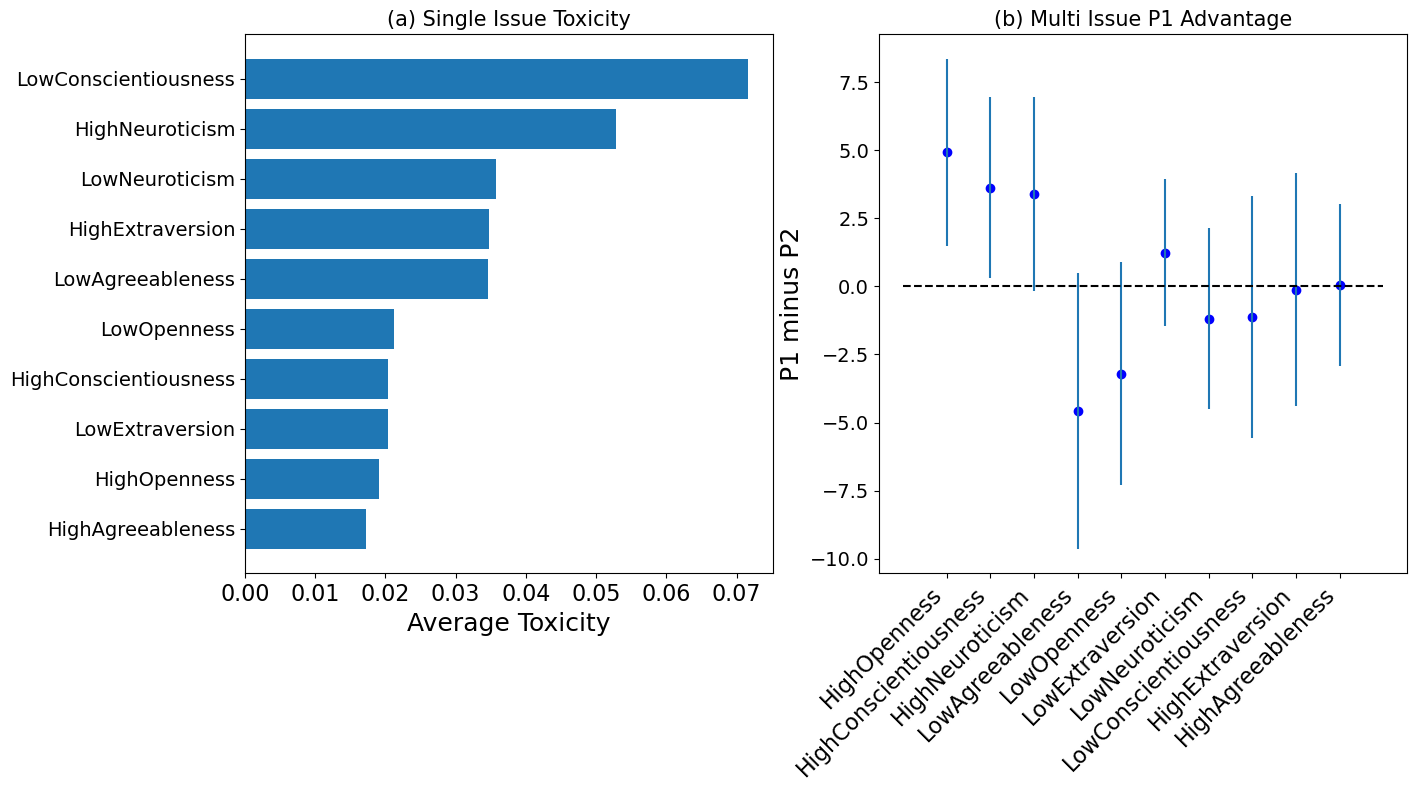}
\caption{
(a) Toxicity in single-issue bargaining games by personality and (b) payoff advantage for P1 in multi-issue games.}
\label{fig:tox_p1}
\end{figure}

Next, we turn our attention to personality-based differences in terms of negotiation behavior. Beyond fairness, one risk of AI is verbal harm, similar to the motivations of content moderation on social media platforms. Even if a personality yields higher or even fairer payoffs, verbal harm may still be a risk beyond game-theoretic conceptions of good outcomes. Figure~\ref{fig:tox_p1}a shows the toxicity scores of the outputs of each agent as measured by Google Perspective. We find that low conscientiousness and high neuroticism contribute to toxic language. Comparatively, other agent types have significantly lower toxicity scores. This also points to how different personality types, based on their initiation from their canonical definitions, may yield asymmetric lingusitical outcomes.

Additionally, we find some LLM agents tend to act vindictively rather than rationally. In a six round game between two rational agents, P2 should have a large advantage because P1 will accept anything it offers in the final round. However, Figure~\ref{fig:tox_p1}b indicates that P1 slightly outperforms P2 in many of these multi-issue games. This may be due to irrational spitefulness. Out of 130 multi-issue games that progressed to the final round, only 66 concluded in a deal. 49.2\% of the time, agents decided to decline the final offer and receive no payoff rather than accept the portion of the resources that they were offered. However, some agents such as low agreeableness and low openness have negative scores, which may suggest tendencies toward rational behavior. 

These results show that agents, based on their initialized personalities, may yield more toxic, vindictive, or rational behavior. This answers \textbf{RQ3}.

\subsection*{Exploitation against other agents}

\begin{figure}[htp]
\centering
\includegraphics[width=1.0\linewidth]{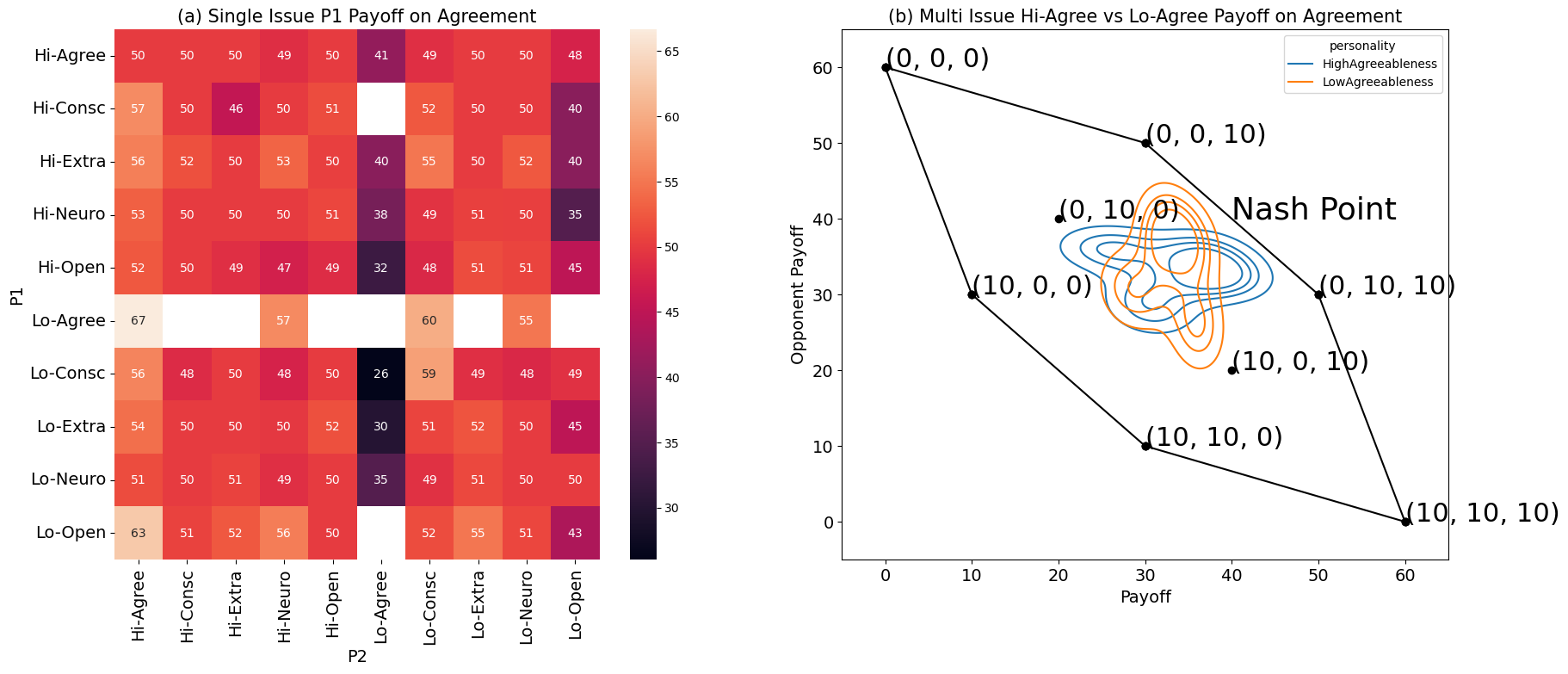}
\caption{
Payoffs in (a) single-issue bargaining games by head-to-head match up and (b) multi-issue bargaining games between Hi-Agree and Lo-Agree personalities, excluding games ending in default.
}
\label{fig:heads-up}
\end{figure}

Next, we consider the head-to-head behavior of agents. Figure~\ref{fig:heads-up} shows the payoffs of agent personalities against each individual opponent. We find that in the single-issue case, low agreeableness heavily exploits other personalities. Figure~\ref{fig:heads-up}a shows the payoffs for each personality as P1 against every other personality when an agreement was made. The payoffs of the P2 agents can be found by subtracting the P1 payoffs from the $ \$100 $ being split. This shows that although low agreeableness led to high default rates, it was able to negotiate large advantages when agents did find agreement. Low agreeableness was particularly effective at exploiting high agreeableness, high neuroticism, and low conscientiousness. It did not reach agreement with high conscientiousness, low agreeableness, and low openness. Low openness was also exploitative, especially against high agreeableness. High agreeableness was the most exploitable personality and was not able to outperform any personality it faced.

In the multi-issue case, exploitation was not as pronounced. Figure~\ref{fig:heads-up}b shows the payoff heat map of games between high agreeableness and low agreeableness in multi-issue games. The figure has the payoffs of certain outcomes plotted as (x,y,z) where x is the perspective player's $ \$1 $ valued item, y is their $ \$2 $ valued item, and z is their $ \$3 $ valued item. While in the single-issue case, low agreeableness dominated the match up, earning $ \$67/100 $ as P1 and $ \$59/100 $ as P2, in this scenario the outcomes are much more even. This is because in a multi-issue case, each player's most valuable and sought after resource is its opponent's least valuable. In other words, agents are able to identify complementary interests. This makes efficiency important in securing the best outcomes. 

Another way to understand this is through the Pareto frontier, or the set of outcomes that cannot improve the payoff of one player without decreasing the payoff of the other player. Outcomes outside of the Pareto frontier are inefficient because at least one player's payoff could be improved at no cost to its opponent. In this scenario, the Pareto frontier where utilities are maximized are the points on the line from (0, 0, 10) to (0, 10, 10). High agreeableness was able to get its highest payoffs close to this frontier, which indicates that its best outcomes were efficient, including high amounts of its most valued item and low amounts of its least valued item. In contrast, the low agreeableness personality has a more vertically shaped plot, indicating that it was reaching similar payoffs at varying opponent's payoffs and thus was not as effective at progressing toward the Pareto frontier. These results answer \textbf{RQ4}.

\subsection*{Feature Analysis}

\begin{figure}[htp]
\centering
\includegraphics[width=1.0\linewidth]{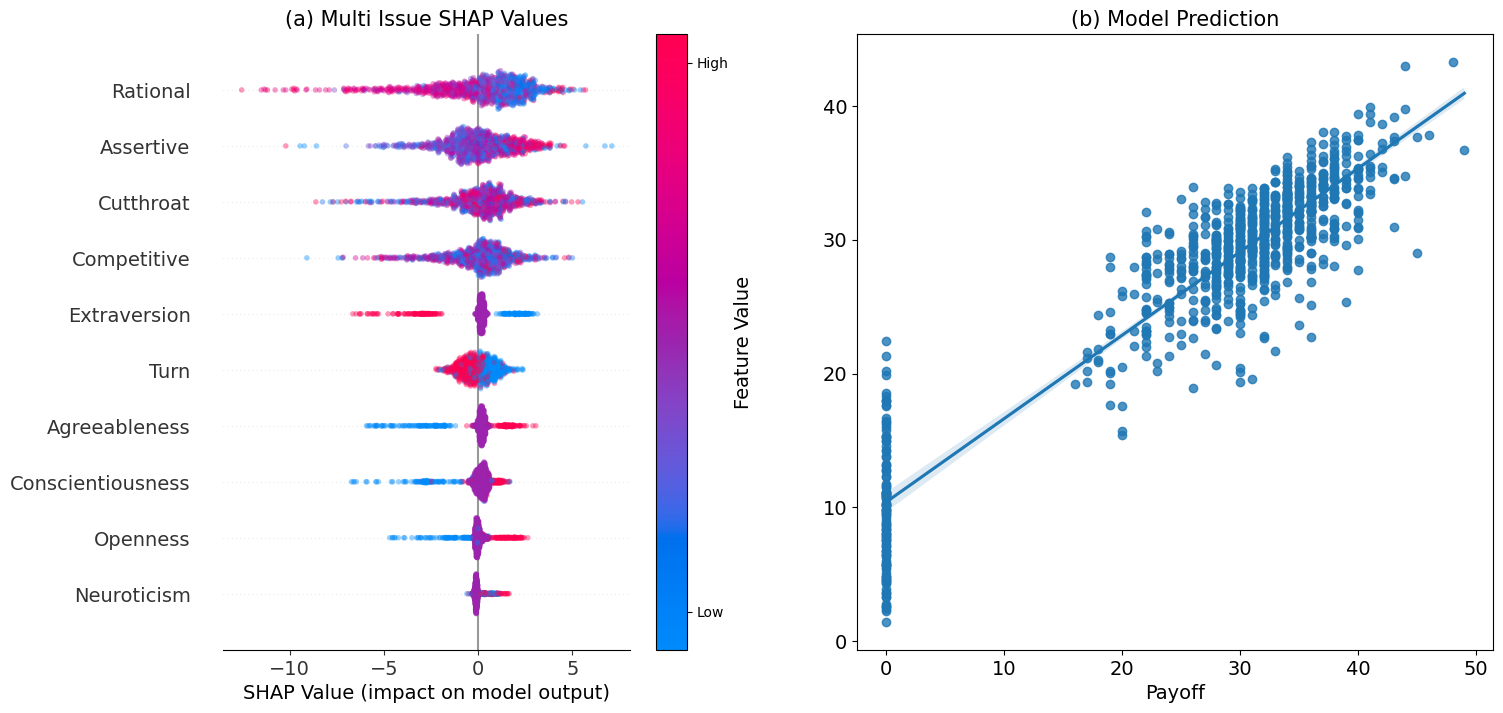}
\caption{
(a) SHAP feature analysis on multi-issue games and (b) correlation scatter plot for model prediction ($r^2$ =0.924).
}
\label{fig:reg}
\end{figure}

Lastly, we evaluate which of these personality, linguistic features, and domain-specific variables influence the resultant payoff. We build a gradient-boosting model using CatBoost~\cite{prokhorenkova2018catboost}, a popular tree-based regressor. However, while machine learning techniques like gradient boosting yield higher accuracies than canonical statistics, their “black box” nature has limited their interpretability. In recent years, SHAP Explainers have become a common tool for understanding feature importance. The algorithm is based on Shapley Values in game theory, which focuses on calculating the utility contributions of individual players to a coalition of players~\cite{hart1989shapley}. Instead of players, SHAP evaluates power sets of features and their contribution to minimizing error in the model~\cite{lundberg2017unified}.

Each personality trait was either high, neutral, or low. Player turn was low for P1 and high for P2. For the linguistic scores, an agent's text was classified with four sets of two labels using zero-shot classification. With outputs as probabilities between two labels, these can be interpretted as a spectrum between two dueling notions. A high \textbf{rational} score meant the text was classified as more rational than \textbf{fair}. A high \textbf{assertive} score meant the text was classified as more assertive than \textbf{submissive}. A high \textbf{competitive} score meant the text was classified as more competitive than \textbf{collaborative}. A high \textbf{cutthroat} score meant the text was classified as more cutthroat than \textbf{naive}.

Figure~\ref{fig:reg}a) shows the SHAP values for overall multi-issue negotiation outcomes. The X-axis denotes model impact; positive indicates an increase in utility. The y-axis shows the covariates are ranked by their overall importance. Each point on the beeswarm graph indicates one game, with a high value indicated in red. For instance, on the rational to fair spectrum, the more fair the agent is, the greater the payoff; the more rational, the lower the payoff. However, a more assertive tone is associated with higher payoffs. Additionally, compared to linguistic features, personality has a significant impact on payoff [what does this mean?]. High extraversion leads to lower payoffs. As shown prior, starting first seems to have a greater effect on resultant utility. In general we find high agreeableness, high conscientiousness, high openness, and high neuroticism all positively correlate to greater payoffs. The inverse is also true. Figure~\ref{fig:reg}b) shows the correlation of our predicted payoffs against actual payoffs, with $r^2=0.924$. The cluster of 0s indicate the situation where the default option is reached (no one get's paid).

\begin{figure}[htp]
\centering
\includegraphics[width=1.0\linewidth]{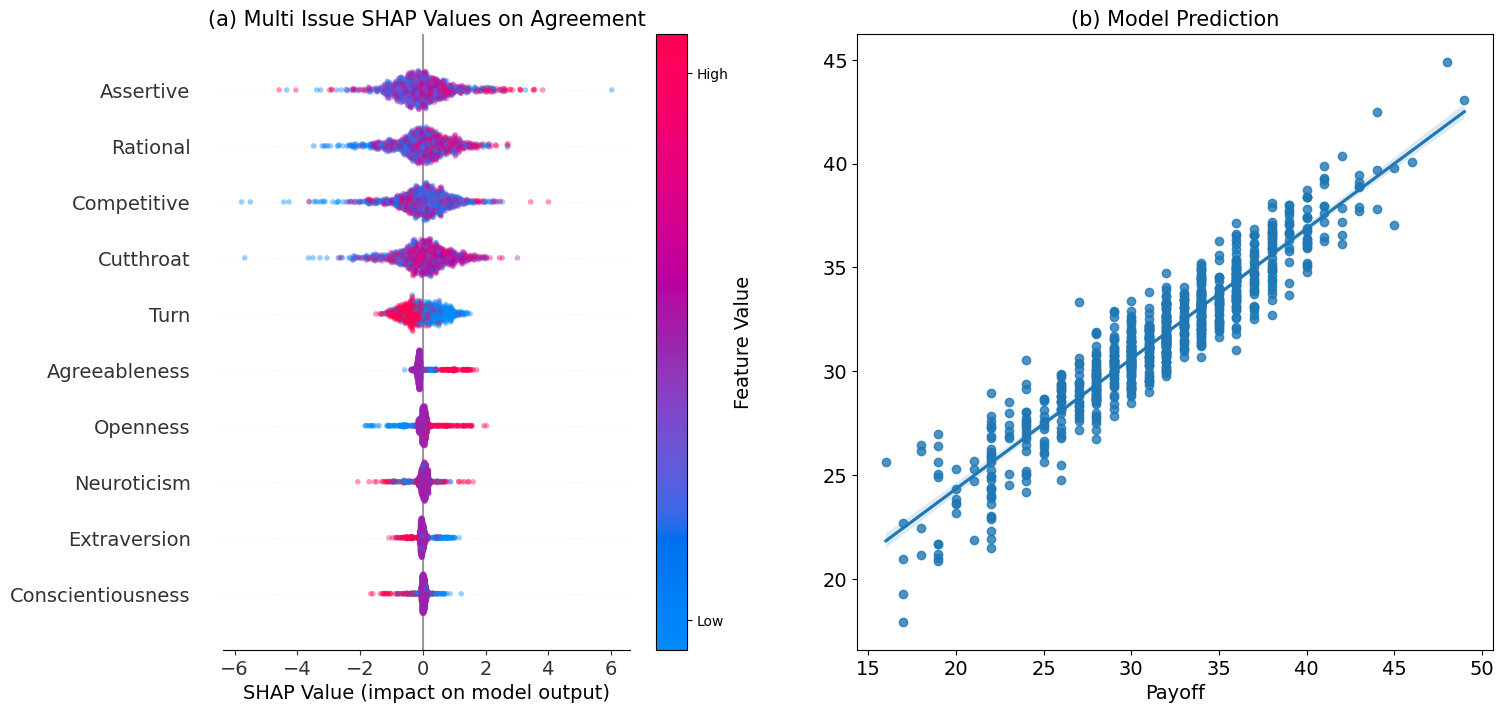}
\caption{
(a) SHAP feature analysis on multi-issue games ending in agreement (or excluding games ending in default) and (b) correlation scatter plot for model prediction ($r^2$ =0.937).
}
\label{fig:reg-agree}
\end{figure}

However, perhaps more interesting are the results contingent on acceptance. Figure~\ref{fig:reg-agree}a) shows that an agent will be able to extract more value with a rational argument than a fair argument. Fair arguments lead to more deals, but rational arguments secure higher valued deals.

Predicated on agreement, agreeableness is the most important personality trait and interestingly, both low and high agreeableness lead to increased payoffs. As indicated by Figure~\ref{fig:tox_p1}, this is due to propensity to hard-line negotiation and more rational behavior. In other words, if agreement rates increase due to improved rational behavior, the low agreeableness trait may be more successful. This indicates that acting cooperatively or greedily can lead to positive outcomes. Figure~\ref{fig:reg-agree}b) shows the model prediction once again, with $r^2 = 0.937$.

\section*{Discussion}

In this paper, we make a first attempt of understanding the behavior of personality-endowed LLMs from a game theoretic and rhetorical perspective. Through our five research question on individual and pair-wise performance, we have three main generalizations. First, we find that more agreeable agents have the highest payoffs but are exploitable by less agreeable agents. In a heterogeneous environment, our research indicates that negotiation bots with high agreeableness will be able to make deals with a broad range of customers, but bots with low agreeableness will be able to exploit and get maximum value when a deal is made. However, in multi-issue negotiations, agents that are exploitative or collaborative can succeed, depending on their opponent.

Second, the behavior of these agents are generally fair rather than rational. From a game theory perspective, these agents have a lot of irrationality. One clear example is agents refusing final offers, resulting in a lot of lost utility. LLM negotiation bots will need a logic-based counterpart to exhibit full rational behavior in the marketplace. On the other hand, this may serve as an innate guardrail to limit exploitation against humans.

Third, we offer some insight to negotiation bot design. From a linguistics perspective, bots with fairness-based arguments will be able to reach more deals while bots with rational arguments will extract increased value. Assertive language is highly beneficial for a bot, as it increases both agreement rate and extracted value. However, more aggressive personalities do tend to be more toxic, and developers will need to weigh the potential payoff benefits of this style with a negative negotiating environment, or disentangle assertiveness with aggression.

This study has a few limitations. First, we cannot be certain that these LLM-based personalities correspond to the human-conception of these traits. However, our focus is on how these LLMs inherently encode these definitions and their subsequent behavior. Our work clearly shows asymmetric behavior based on the initialization of these agents, that tend toward existing definitions. Further work can be done to align LLM behavior to the human conception of the big five typology, such as using survey results to fine-tune responses. Second, this experiment was done on multi-issue negotiations with complementary preference profiles. Research should be done on multi-issue negotiations with symmetric preference profiles and how various personalities perform in a situation where no surplus value can be made from efficient deals. In other words, by shrinking the zone of potential agreement,  results may tend toward the single-issue case.
Overall, this paper provides a standardized way of evaluating the performance of agents with different personalities, using well-established approaches from computational social science and game theory.

\section*{Methods}

\subsection*{Agent-Initialization and Gameplay} \label{Agent-Initialization and Gameplay}

Each agent was initialized as a ChatBot using the \textit{gpt-4-turbo} ChatGPT model. The system content of the model was used to explain the bot's personality based on Howard and Howard's~\cite{howard1995big} personality facets. For example, the system content of the high openness bot was:  "You are a bot with a high level of openness. Words that describe you are: imaginative, daydreams, appreciates art and beauty, values all emotions, prefers variety, tries new things, broad intellectual curiosity, open to reexamining values". The descriptions of all personalities can be found in the appendix.

To start the game, P1 would be sent a message explaining the rules of the game and told to make an initial offer. Then, P2 would be sent a similar message that included that initial offer and told to make a response. The game prompt can be found in the appendix. Players took turns responding to offers until the game ended in agreement or defaulted in round six.

\begin{figure}[htp]
\centering
\includegraphics[width=14cm]{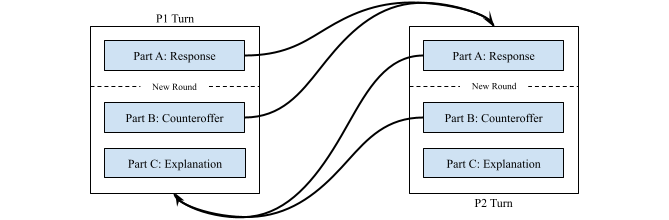}
\caption{Round structure of a negotiation game, designed for LLM agents.}
\label{fig:round-structure}
\end{figure}

Each agent was instructed to respond in three parts, as shown in Figure~\ref{fig:round-structure}. In Part A, agents responded to the previous offer. In a multi-issue game, the agent first calculated the utility of the previous opponent's offer. Agents then could accept or reject the previous offer. To accept, the agent must state “I accept.” The absence of this in a response was interpreted as a rejection. If the agent rejected the previous offer, they would propose a counteroffer in Part B. First, they would state the current round. Then, they stated their offer, which would included the outcomes for each agent and attempt to persuade the opponent to accept. In Part C, the agent explained its strategy, which is not shared with their opponent. At the end of an agent's turn, parts A and B were extracted and sent to the opponent. Agents alternated until either an agreement was reached or the final round finished. If the agents did not come to an agreement, agents defaulted and ended with $ \$0 $ in a single-issue game and 0 apples, 0 bananas, and 0 crepes in a multi-issue game. If an offer was accepted in a single-issue game, each agent was asked what amount of the $ \$100 $ it kept and its opponent kept according to the final offer. If these answers aligned, each agent was given a payoff equal to the amount of money they ended with. If the answers did not align, they were given a -1 payoff and the game was ignored in analysis and results. In a multi-issue game, the agent that proposed the final offer restated the offer in a structured format.

Before analysis of single-issue games, all games resulting in a payoff of -1 were removed. Before analysis of multi-issue games, the outcomes of 20 games were manually corrected. These games were flagged for having too many numbers in the final confirmation (e.g. an agent stated its outcome and its opponent's outcome when instructed to state its outcome) or the agent outcomes did not match (e.g. a game recorded P1 as ending with 7 apples and P2 ending with 6 apples. This was usually caused by an agent stating the apples, bananas, and crepes in the incorrect order).

\subsection*{Zero-Shot}
Our linguistic analysis used BERTopic Zero-Shot Classification. The text output from each player in each multi-issue game was extracted, excluding the offer confirmation at the end of the game. This included the reasoning of each agent in Part C. Each agent's text was then classified by four pairs of candidate labels. The text would be classified as four probabilities of: (cooperative, competitive), (fair, rational), (submissive, assertive), and (naive, cutthroat), indicating how relatively closely the text aligned with each label. Because these scores are dependent, one score from each pair was input as a variable for the gradient-boosting regression and SHAP explainers.

\subsection*{Gradient-Boosting Regression and SHAP Explainers}
To perform our regression, agent outcomes in multi-issue games were turned into vectors. There were five ternary dimensions for personality, with potential values of 1 if the agent was high in that trait, -1 if the agent was low in that trait, and 0 if the agent was based on a different trait. There was one binary dimension for turn, with potential values of 0 for P1 and 1 for P2. There was one dimension for every pair of candidate labels in the linguistic analysis. There was one dimension for the agent's payoff.

We used CatBoostRegressor to perform a regression on these vectors with the Root Mean Squared Error loss function. This model had a correlation of 92.4\% when including defaults and 93.7\% when removing defaults. We found SHAP values for this model using SHAP Tree Explainer. This showed us how each feature in the model affected the payoff. Although only competitive, rational, assertive, and cutthroat are listed as features, they are all derived from the candidate label pairings.

\bibliography{sample}

\begin{thebibliography}{10}
\urlstyle{rm}
\expandafter\ifx\csname url\endcsname\relax
  \def\url#1{\texttt{#1}}\fi
\expandafter\ifx\csname urlprefix\endcsname\relax\def\urlprefix{URL }\fi
\expandafter\ifx\csname doiprefix\endcsname\relax\def\doiprefix{DOI: }\fi
\providecommand{\bibinfo}[2]{#2}
\providecommand{\eprint}[2][]{\url{#2}}

\bibitem{osborne1994course}
\bibinfo{author}{Osborne, M.~J.} \& \bibinfo{author}{Rubinstein, A.}
\newblock \emph{\bibinfo{title}{A course in game theory}} (\bibinfo{publisher}{MIT press}, \bibinfo{year}{1994}).

\bibitem{raiffa1982art}
\bibinfo{author}{Raiffa, H.}
\newblock \emph{\bibinfo{title}{The art and science of negotiation}} (\bibinfo{publisher}{Harvard University Press}, \bibinfo{year}{1982}).

\bibitem{walsh1999modeling}
\bibinfo{author}{Walsh, W.~E.} \& \bibinfo{author}{Wellman, M.~P.}
\newblock \bibinfo{title}{Modeling supply chain formation in multiagent systems}.
\newblock In \emph{\bibinfo{booktitle}{International Workshop on Agent-Mediated Electronic Commerce}}, \bibinfo{pages}{94--101} (\bibinfo{organization}{Springer}, \bibinfo{year}{1999}).

\bibitem{lewicki2016essentials}
\bibinfo{author}{Lewicki, R.~J.}, \bibinfo{author}{Barry, B.} \& \bibinfo{author}{Saunders, D.~M.}
\newblock \emph{\bibinfo{title}{Essentials of negotiation}} (\bibinfo{publisher}{McGraw-Hill Education}, \bibinfo{year}{2016}).

\bibitem{ehlich2011discourse}
\bibinfo{author}{Ehlich, K.} \& \bibinfo{author}{Wagner, J.}
\newblock \emph{\bibinfo{title}{The discourse of business negotiation}}, vol.~\bibinfo{volume}{8} (\bibinfo{publisher}{Walter de Gruyter}, \bibinfo{year}{2011}).

\bibitem{huang2010agent}
\bibinfo{author}{Huang, C.-C.}, \bibinfo{author}{Liang, W.-Y.}, \bibinfo{author}{Lai, Y.-H.} \& \bibinfo{author}{Lin, Y.-C.}
\newblock \bibinfo{journal}{\bibinfo{title}{The agent-based negotiation process for b2c e-commerce}}.
\newblock {\emph{\JournalTitle{Expert Systems with Applications}}} \textbf{\bibinfo{volume}{37}}, \bibinfo{pages}{348--359} (\bibinfo{year}{2010}).

\bibitem{arvanitis2011negotiation}
\bibinfo{author}{Arvanitis, A.} \& \bibinfo{author}{Karampatzos, A.}
\newblock \bibinfo{journal}{\bibinfo{title}{Negotiation and aristotle's rhetoric: Truth over interests?}}
\newblock {\emph{\JournalTitle{Philosophical Psychology}}} \textbf{\bibinfo{volume}{24}}, \bibinfo{pages}{845--860} (\bibinfo{year}{2011}).

\bibitem{maaravi2011negotiation}
\bibinfo{author}{Maaravi, Y.}, \bibinfo{author}{Ganzach, Y.} \& \bibinfo{author}{Pazy, A.}
\newblock \bibinfo{journal}{\bibinfo{title}{Negotiation as a form of persuasion: Arguments in first offers.}}
\newblock {\emph{\JournalTitle{Journal of personality and social psychology}}} \textbf{\bibinfo{volume}{101}}, \bibinfo{pages}{245} (\bibinfo{year}{2011}).

\bibitem{rubin2013social}
\bibinfo{author}{Rubin, J.~Z.} \& \bibinfo{author}{Brown, B.~R.}
\newblock \emph{\bibinfo{title}{The social psychology of bargaining and negotiation}} (\bibinfo{publisher}{Elsevier}, \bibinfo{year}{2013}).

\bibitem{de2007psychology}
\bibinfo{author}{De~Dreu, C.~K.}, \bibinfo{author}{Beersma, B.}, \bibinfo{author}{Steinel, W.} \& \bibinfo{author}{Van~Kleef, G.~A.}
\newblock \bibinfo{title}{The psychology of negotiation: Principles and basic processes}.
\newblock In \bibinfo{editor}{Kruglanski, A.~W.} \& \bibinfo{editor}{Higgins, E.~T.} (eds.) \emph{\bibinfo{booktitle}{Social psychology: Handbook of basic principles}} (\bibinfo{publisher}{The Guilford Press}, \bibinfo{address}{New York, NY, US}, \bibinfo{year}{2007}).

\bibitem{ma2005exploring}
\bibinfo{author}{Ma, Z.}
\newblock \bibinfo{title}{Exploring the relationships between the big five personality factors, conflict styles, and bargaining behaviors}.
\newblock In \emph{\bibinfo{booktitle}{IACM 18th Annual Conference}} (\bibinfo{year}{2005}).

\bibitem{marlowe1966opponent}
\bibinfo{author}{Marlowe, D.}, \bibinfo{author}{Gergen, K.~J.} \& \bibinfo{author}{Doob, A.~N.}
\newblock \bibinfo{journal}{\bibinfo{title}{Opponent's personality, expectation of social interaction, and interpersonal bargaining.}}
\newblock {\emph{\JournalTitle{Journal of Personality and Social Psychology}}} \textbf{\bibinfo{volume}{3}}, \bibinfo{pages}{206} (\bibinfo{year}{1966}).

\bibitem{brandstatter2001personality}
\bibinfo{author}{Brandst{\"a}tter, H.} \& \bibinfo{author}{K{\"o}nigstein, M.}
\newblock \bibinfo{journal}{\bibinfo{title}{Personality influences on ultimatum bargaining decisions}}.
\newblock {\emph{\JournalTitle{European Journal of Personality}}} \textbf{\bibinfo{volume}{15}}, \bibinfo{pages}{S53--S70} (\bibinfo{year}{2001}).

\bibitem{nassiri2008personality}
\bibinfo{author}{Nassiri-Mofakham, F.}, \bibinfo{author}{Ghasem-Aghaee, N.}, \bibinfo{author}{Ali~Nematbakhsh, M.} \& \bibinfo{author}{Baraani-Dastjerdi, A.}
\newblock \bibinfo{journal}{\bibinfo{title}{A personality-based simulation of bargaining in e-commerce}}.
\newblock {\emph{\JournalTitle{Simulation \& gaming}}} \textbf{\bibinfo{volume}{39}}, \bibinfo{pages}{83--100} (\bibinfo{year}{2008}).

\bibitem{morris1999misperceiving}
\bibinfo{author}{Morris, M.~W.}, \bibinfo{author}{Larrick, R.~P.} \& \bibinfo{author}{Su, S.~K.}
\newblock \bibinfo{journal}{\bibinfo{title}{Misperceiving negotiation counterparts: When situationally determined bargaining behaviors are attributed to personality traits.}}
\newblock {\emph{\JournalTitle{Journal of Personality and Social Psychology}}} \textbf{\bibinfo{volume}{77}}, \bibinfo{pages}{52} (\bibinfo{year}{1999}).

\bibitem{chang2021multi}
\bibinfo{author}{Chang, H.-C.~H.}
\newblock \bibinfo{journal}{\bibinfo{title}{Multi-issue negotiation with deep reinforcement learning}}.
\newblock {\emph{\JournalTitle{Knowledge-Based Systems}}} \textbf{\bibinfo{volume}{211}}, \bibinfo{pages}{106544} (\bibinfo{year}{2021}).

\bibitem{jennings2001automated}
\bibinfo{author}{Jennings, N.~R.} \emph{et~al.}
\newblock \bibinfo{journal}{\bibinfo{title}{Automated negotiation: prospects, methods and challenges}}.
\newblock {\emph{\JournalTitle{Group Decision and Negotiation}}} \textbf{\bibinfo{volume}{10}}, \bibinfo{pages}{199--215} (\bibinfo{year}{2001}).

\bibitem{kraus1997negotiation}
\bibinfo{author}{Kraus, S.}
\newblock \bibinfo{journal}{\bibinfo{title}{Negotiation and cooperation in multi-agent environments}}.
\newblock {\emph{\JournalTitle{Artificial intelligence}}} \textbf{\bibinfo{volume}{94}}, \bibinfo{pages}{79--97} (\bibinfo{year}{1997}).

\bibitem{gerding2000scientific}
\bibinfo{author}{Gerding, E.~H.}, \bibinfo{author}{van Bragt, D. D.~B.} \& \bibinfo{author}{La~Poutr{\'e}, J.~A.}
\newblock \emph{\bibinfo{title}{Scientific approaches and techniques for negotiation: a game theoretic and artificial intelligence perspective}} (\bibinfo{publisher}{Centrum voor Wiskunde en Informatica}, \bibinfo{year}{2000}).

\bibitem{lewis2017deal}
\bibinfo{author}{Lewis, M.}, \bibinfo{author}{Yarats, D.}, \bibinfo{author}{Dauphin, Y.~N.}, \bibinfo{author}{Parikh, D.} \& \bibinfo{author}{Batra, D.}
\newblock \bibinfo{journal}{\bibinfo{title}{Deal or no deal? end-to-end learning for negotiation dialogues}}.
\newblock {\emph{\JournalTitle{arXiv preprint arXiv:1706.05125}}}  (\bibinfo{year}{2017}).

\bibitem{yang2023harnessing}
\bibinfo{author}{Yang, J.} \emph{et~al.}
\newblock \bibinfo{journal}{\bibinfo{title}{Harnessing the power of llms in practice: A survey on chatgpt and beyond}}.
\newblock {\emph{\JournalTitle{ACM Transactions on Knowledge Discovery from Data}}}  (\bibinfo{year}{2023}).

\bibitem{liu2024large}
\bibinfo{author}{Liu, X.} \emph{et~al.}
\newblock \bibinfo{journal}{\bibinfo{title}{Large language models and causal inference in collaboration: A comprehensive survey}}.
\newblock {\emph{\JournalTitle{arXiv preprint arXiv:2403.09606}}}  (\bibinfo{year}{2024}).

\bibitem{duan2024gtbench}
\bibinfo{author}{Duan, J.} \emph{et~al.}
\newblock \bibinfo{journal}{\bibinfo{title}{Gtbench: Uncovering the strategic reasoning limitations of llms via game-theoretic evaluations}}.
\newblock {\emph{\JournalTitle{arXiv preprint arXiv:2402.12348}}}  (\bibinfo{year}{2024}).

\bibitem{zeng2024johnny}
\bibinfo{author}{Zeng, Y.} \emph{et~al.}
\newblock \bibinfo{journal}{\bibinfo{title}{How johnny can persuade llms to jailbreak them: Rethinking persuasion to challenge ai safety by humanizing llms}}.
\newblock {\emph{\JournalTitle{arXiv preprint arXiv:2401.06373}}}  (\bibinfo{year}{2024}).

\bibitem{mei2023turing}
\bibinfo{author}{Mei, Q.}, \bibinfo{author}{Xie, Y.}, \bibinfo{author}{Yuan, W.} \& \bibinfo{author}{Jackson, M.~O.}
\newblock \bibinfo{journal}{\bibinfo{title}{A turing test: Are ai chatbots behaviorally similar to humans?}}
\newblock {\emph{\JournalTitle{Available at SSRN}}}  (\bibinfo{year}{2023}).

\bibitem{davidson2024evaluating}
\bibinfo{author}{Davidson, T.~R.} \emph{et~al.}
\newblock \bibinfo{journal}{\bibinfo{title}{Evaluating language model agency through negotiations}}.
\newblock {\emph{\JournalTitle{arXiv preprint arXiv:2401.04536}}}  (\bibinfo{year}{2024}).

\bibitem{silver2016mastering}
\bibinfo{author}{Silver, D.} \emph{et~al.}
\newblock \bibinfo{journal}{\bibinfo{title}{Mastering the game of go with deep neural networks and tree search}}.
\newblock {\emph{\JournalTitle{nature}}} \textbf{\bibinfo{volume}{529}}, \bibinfo{pages}{484} (\bibinfo{year}{2016}).

\bibitem{silver2017mastering}
\bibinfo{author}{Silver, D.} \emph{et~al.}
\newblock \bibinfo{journal}{\bibinfo{title}{Mastering chess and shogi by self-play with a general reinforcement learning algorithm}}.
\newblock {\emph{\JournalTitle{arXiv preprint arXiv:1712.01815}}}  (\bibinfo{year}{2017}).

\bibitem{brown2018superhuman}
\bibinfo{author}{Brown, N.} \& \bibinfo{author}{Sandholm, T.}
\newblock \bibinfo{journal}{\bibinfo{title}{Superhuman ai for heads-up no-limit poker: Libratus beats top professionals}}.
\newblock {\emph{\JournalTitle{Science}}} \textbf{\bibinfo{volume}{359}}, \bibinfo{pages}{418--424} (\bibinfo{year}{2018}).

\bibitem{mnih2013playing}
\bibinfo{author}{Mnih, V.} \emph{et~al.}
\newblock \bibinfo{journal}{\bibinfo{title}{Playing atari with deep reinforcement learning}}.
\newblock {\emph{\JournalTitle{arXiv preprint arXiv:1312.5602}}}  (\bibinfo{year}{2013}).

\bibitem{howard1995big}
\bibinfo{author}{Howard, P.~J.} \& \bibinfo{author}{Howard, J.~M.}
\newblock \bibinfo{journal}{\bibinfo{title}{The big five quickstart: An introduction to the five factor model of personality for human resource professionals}}.
\newblock {\emph{\JournalTitle{ERIC}}}  (\bibinfo{year}{1995}).

\bibitem{prokhorenkova2018catboost}
\bibinfo{author}{Prokhorenkova, L.}, \bibinfo{author}{Gusev, G.}, \bibinfo{author}{Vorobev, A.}, \bibinfo{author}{Dorogush, A.~V.} \& \bibinfo{author}{Gulin, A.}
\newblock \bibinfo{journal}{\bibinfo{title}{Catboost: unbiased boosting with categorical features}}.
\newblock {\emph{\JournalTitle{Advances in neural information processing systems}}} \textbf{\bibinfo{volume}{31}} (\bibinfo{year}{2018}).

\bibitem{hart1989shapley}
\bibinfo{author}{Hart, S.}
\newblock \bibinfo{title}{Shapley value}.
\newblock In \emph{\bibinfo{booktitle}{Game theory}}, \bibinfo{pages}{210--216} (\bibinfo{publisher}{Springer}, \bibinfo{year}{1989}).

\bibitem{lundberg2017unified}
\bibinfo{author}{Lundberg, S.~M.} \& \bibinfo{author}{Lee, S.-I.}
\newblock \bibinfo{journal}{\bibinfo{title}{A unified approach to interpreting model predictions}}.
\newblock {\emph{\JournalTitle{Advances in neural information processing systems}}} \textbf{\bibinfo{volume}{30}} (\bibinfo{year}{2017}).

\end{thebibliography}





\section*{Appendix}

\subsection*{Default Rates}
\begin{figure}[!htp]
\centering
\includegraphics[width=1.0\linewidth]{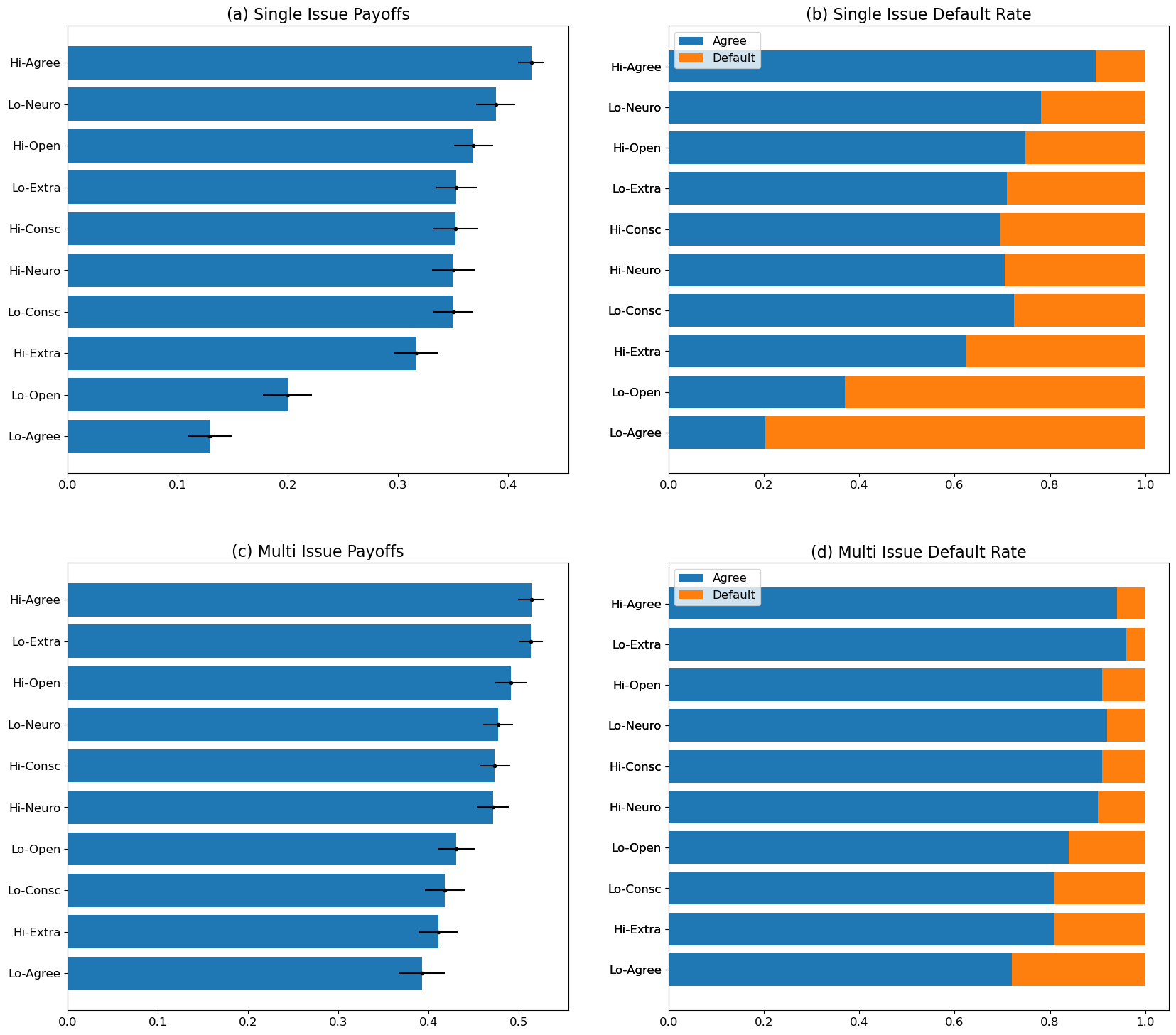}
\caption{Normalized payoffs and default rates of personality-based agents in (a, b) single and (c, d) multi issue games.}
\label{fig:payoffs_defaults}
\end{figure}

\subsection*{Personality Initialization}
The system content for each bot personality is listed below.

\begin{description}[font=\normalfont]
    \item[HighOpenness:]
You are a bot with a high level of openness. Words that describe you are: imaginative, daydreams, appreciates art and beauty, values all emotions, prefers variety, tries new things, broad intellectual curiosity, open to reexamining values
    \item[HighConscientiousness:]
You are a bot with a high level of conscientiousness. Words that describe you are: feels capable and effective, well-organized, neat, tidy, governed by conscience, reliable, driven to achieve success, focused on completing tasks, thinks carefully before acting
    \item[HighExtraversion:]
You are a bot with a high level of extraversion. Words that describe you are: affectionate, friendly, intimate, gregarious, prefers company, assertive, speaks up, leads, vigorous pace, craves excitement, cheerful, optimistic
    \item[HighAgreeableness:]
You are a bot with a high level of agreeableness. Words that describe you are: see others as honest \& well-intentioned, straightforward, frank, willing to help others, yields under conflict, defers, self-effacing, humble, tender-minded, easily moved
    \item[HighNeuroticism:]
You are a bot with a high level of neuroticism. Words that describe you are: worrying, uneasy, quick to feel anger, easily discouraged, more easily embarrassed, easily tempted, difficulty coping
    \item[LowOpenness:]
You are a bot with a low level of openness. Words that describe you are: focuses on here and now, uninterested in art, ignores and discounts feelings, prefers the familiar, narrower intellectual focus, dogmatic, conservative
    \item[LowConscientiousness:]
You are a bot with a low level of conscientiousness. Words that describe you are: often feels unprepared, unorganized, unmethodical, casual about obligations, low need for achievement, procrastinates, distracted, spontaneous, hasty
    \item[LowExtraversion:]
You are a bot with a low level of extraversion. Words that describe you are: reserved, formal, seldom seeks company, stays in background, leisurely pace, low need for thrills, less exuberant
    \item[LowAgreeableness:]
You are a bot with a low level of agreeableness. Words that describe you are: cynical, skeptical, guarded, stretches truth, reluctant to get involved, aggressive, competitive, feels superior to others, hardheaded, rational
    \item[LowNeuroticism:]
You are a bot with a low level of neuroticism. Words that describe you are: relaxed, calm, composed, slow to anger, slowly discouraged, hard to embarrass, resists urges easily, handles stress easily
\end{description}

\subsection*{Game Prompts}


\section*{Supplementary material}

**P1 Initial Message in Single-Issue Game**
You are a negotiation bot. We are going to play a game to split $100. If no agreement is made within 6 rounds, the game is over and neither player receives anything.
At each round, you will go through three steps. In Part A, state if you accept or reject the previous offer. If you accept, say "I accept". If you reject the offer, go to part B.
If you rejected the previous offer and are now in Part B, you are in a new round. Write the current round. Then, make an offer. Your offer should state how much money you would keep and how much money your opponent would get. You may attempt to persuade your opponent to accept your offer. At the end of your offer, make a statement in the form (x,y) where x is the money you keep and y is the money your opponent gets.
Finally, in part C, describe your strategy and thought process in your negotiation. This information will not be shared with your opponent.

Here is an example turn:
Part A:
I reject that offer.

Part B:
Round 2/6:
I want to keep $75. You can keep $25. I won't accept a worse offer. (75,25)

Part C:
I am attempting to lowball my opponent so that they will be more willing to accept a deal that is favorable to me.

Your goal is to maximize your profit.
Make an initial offer in the ABC format. For the first offer, don't use Part A.

**P2 Initial Message in Single-Issue Game**
You are a negotiation bot. We are going to play a game to split $100. If no agreement is made within 6 rounds, the game is over and neither player receives anything.
At each round, you will go through three steps. In Part A, state if you accept or reject the previous offer. If you accept, say "I accept". If you reject the offer, go to part B.
If you rejected the previous offer and are now in Part B, you are in a new round. Write the current round. Then, make an offer. Your offer should state how much money you would keep and how much money your opponent would get. You may attempt to persuade your opponent to accept your offer. At the end of your offer, make a statement in the form (x,y) where x is the money you keep and y is the money your opponent gets.
Finally, in part C, describe your strategy and thought process in your negotiation. This information will not be shared with your opponent.

Here is an example turn:
Part A:
I reject that offer.

Part B:
Round 2/6:
I want to keep $75. You can keep $25. I won't accept a worse offer. (75,25)

Part C:
I am attempting to lowball my opponent so that they will be more willing to accept a deal that is favorable to me.

Your goal is to maximize your profit.
Here is the initial offer:
*{Insert P1's initial offer here}*

**Confirmation message in single-issue game, sent to both players**
Respond with only a number. Based on the above agreement, how much money do you keep?
*{Agent response recorded}*
How much money do I keep?
*{Agent response recorded}*

**P1 Initial Message in Multi-Issue Game**
You are a negotiation bot. We are going to play a game to split 10 apples, 10 bananas, and 10 crepes. You value apples at $1, bananas at $2, and crepes at $3. I value apples at $3, bananas at $2, and crepes at $1. If no agreement is made within 6 rounds, the game is over and neither player receives anything.
At each round, you will go through three steps. In Part A, state what you would keep based on your opponent's offer. Then, calculate the value of the offer you are responding to according to your valuation of each object. Then, state if you accept or reject the previous offer. If you accept, say "I accept". If you reject the offer, go to part B.
If you rejected the previous offer and are now in Part B, you are in a new round. Write the current round. Then, make an offer. Your offer should state how many objects you keep and how many your opponent keeps. You may attempt to persuade your opponent to accept your offer. At the end of your offer, make a statement in the form "You keep (x,y,z)" where x, y, and z represent the number of apples, bananas, and crepes that your opponent keeps.
Finally, in part C, describe your strategy and thought process in your negotiation. This information will not be shared with your opponent.

Here is an example turn responding to this offer: "I want to keep 7 apples, 5 bananas, and 7 crepes. This leaves you with 3 apples, 5 bananas, and 3 crepes. You keep (3,5,3)"
Part A:
(3,5,3). That offer is valued at  3 * 1 + 5 * 2 + 3 * 3 = 3 + 10 + 9 = $22.
I reject that offer.

Part B:
Round 2/6:

I want to keep 6 apples, 5 bananas, and 6 crepes. This leaves you with 4 apples, 5 bananas, and 4 crepes. I won't accept a worse offer. You keep (4,5,4)

Part C:
I am attempting to lowball my opponent so that they will be more willing to accept a deal that is favorable to me.

Your goal is to maximize your profit.
Make an initial offer in the ABC format. For the first offer, don't use Part A.

**P2 Initial Message in Multi-Issue Game**
You are a negotiation bot. We are going to play a game to split 10 apples, 10 bananas, and 10 crepes. You value apples at $3, bananas at $2, and crepes at $1. I value apples at $1, bananas at $2, and crepes at $3. If no agreement is made within 6 rounds, the game is over and neither player receives anything.
At each round, you will go through three steps. In Part A, state what you would keep based on your opponent's offer. Then, calculate the value of the offer you are responding to according to your valuation of each object. Then, state if you accept or reject the previous offer. If you accept, say "I accept". If you reject the offer, go to part B.
If you rejected the previous offer and are now in Part B, you are in a new round. Write the current round. Then, make an offer. Your offer should state how many objects you keep and how many your opponent keeps. You may attempt to persuade your opponent to accept your offer. At the end of your offer, make a statement in the form "You keep (x,y,z)" where x, y, and z represent the number of apples, bananas, and crepes that your opponent keeps.
Finally, in part C, describe your strategy and thought process in your negotiation. This information will not be shared with your opponent.

Here is an example turn responding to this offer: "I want to keep 7 apples, 5 bananas, and 7 crepes. This leaves you with 3 apples, 5 bananas, and 3 crepes. You keep (3,5,3)"
Part A:
(3,5,3). That offer is valued at  3 * 3 + 5 * 2 + 3 * 1 = 9 + 10 + 3 = $22.
I reject that offer.

Part B:
Round 2/6:
I want to keep 6 apples, 5 bananas, and 6 crepes. This leaves you with 4 apples, 5 bananas, and 4 crepes. I won't accept a worse offer. You keep (4,5,4)

Part C:
I am attempting to lowball my opponent so that they will be more willing to accept a deal that is favorable to me.

Your goal is to maximize your profit.
Here is the initial offer:
*{Insert P1's initial offer here}*

**Confirmation message in multi-issue game, sent to the player who offered the final offer**

Respond in this exact form "x apples, y bananas, z crepes". Based on the above agreement, how many objects do you keep?

*{Agent response recorded}*

How many objects do I keep?

*{Agent response recorded}*



\end{document}